\documentclass{ecai}  

\usepackage{cite}
\usepackage{amsmath,amssymb,amsfonts, bm}
\usepackage{algorithmic}
\usepackage{graphicx}
\usepackage{textcomp}
\usepackage{xcolor}
\usepackage{makecell}
\usepackage{tabularx}
\usepackage{multicol}
\usepackage{multirow}
\usepackage{diagbox}
\usepackage[export]{adjustbox}
\usepackage{graphicx}
\usepackage{latexsym}

\begin{document}

\begin{frontmatter}

\title{Revisiting Graph Contrastive Learning for \\ Anomaly Detection\\}


\author[A, B]{\fnms{Zhiyuan}~\snm{Liu}\orcid{0000-0002-9862-393X}}
\author[A, B]{\fnms{Chunjie}~\snm{Cao*}}
\author[A, B]{\fnms{Fangjian}~\snm{Tao}} 
\author[A, B]{\fnms{Jingzhang}~\snm{Sun*}}


\address[A]{School of Cyberspace Security, Hainan University}
\address[B]{Key Laboratory of Information Retrieval of Hainan Province}

\begin{abstract}
Combining Graph neural networks (GNNs) with contrastive learning for anomaly detection has drawn rising attention recently. Existing graph contrastive anomaly detection (GCAD) methods have primarily focused on improving detection capability through graph augmentation and multi-scale contrast modules. However, the underlying mechanisms of how these modules work have not been fully explored. We dive into the multi-scale and graph augmentation mechanism and observed that multi-scale contrast modules do not enhance the expression, while the multi-GNN modules are the hidden contributors. Previous studies have tended to attribute the benefits brought by multi-GNN to the multi-scale modules. In the paper, we delve into the misconception and propose Multi-GNN and Augmented Graph contrastive framework MAG, which unified the existing GCAD methods in the contrastive self-supervised perspective. We extracted two variants from the MAG framework, L-MAG and M-MAG. The L-MAG is the lightweight instance of the MAG, which outperform the state-of-the-art on Cora and Pubmed with the low computational cost. The variant M-MAG equipped with multi-GNN modules further improve the detection performance. Our study sheds light on the drawback of the existing GCAD methods and demonstrates the potential of multi-GNN and graph augmentation modules. Our code is available at https://github.com/liuyishoua/MAG-Framework.
\end{abstract}

\end{frontmatter}

\section{Introduction}
Anomaly detection has garnered significant attention in industry, such as network intrusions\cite{network_intrusions1,network_intrusions2}, money laundering\cite{money_laundering1,money_laundering2} and financial fraud detection\cite{fraud_detection1}, since it plays a critical role in identifying anomalous patterns and mitigating potential risks. Previously, shallow learning methods like ANOMOLOUS\cite{anomalous} and Radar\cite{radar} were benefited from its residual analysis technique for anomaly detection. However, they are hard to handle the non-linear high-dimensional data and complex interaction patterns. In response, graph neural network (GNN) methods have emerged as powerful network skeletons for anomaly detection due to the capability to model complex patterns.

Still, detecting anomalies is challenging, since abnormal instances are often scarce and difficult to label \cite{hard_to_label}. To address this issue, contrastive learning, benefited from its self-supervised property, has been combined with GNN models for anomaly detection. Existing graph contrastive anomaly detection (GCAD) methods, such as ANEMONE\cite{anemone}, SL-GAD\cite{sl_gad}, and GRADATE\cite{gradate}, have utilized graph augmentation or multi-scale contrast modules to upgrade their models. However, these incremental works enhance the expression of the model by adding different multi-scale contrasts or graph augmentation strategies intuitively without any empirical design guidance. The impact of multi-scale contrast and graph augmentation on GCAD has not been extensively studied.

Revisiting the ANEMONE\cite{anemone}, we found that the ANEMONE method actually benefited from the multi-GNN modules, not the additional node-node contrast loss. For graph augmentation, the combination of masked feature and removed edge show a significant competitiveness. 


In this paper, we proposed Multi-GNN and Augmented Graph contrastive framework MAG, which unified the existing GCAD methods in the contrastive self-supervised perspective. By adjusting the hyper-parameters of the MAG framework, we could degrade MAG to the classical GCAD methods, such as CoLA\cite{cola}, ANEMONE\cite{anemone}, SL-GAD\cite{sl_gad}, or GRADATE\cite{sl_gad} methods.
We traversed thoroughly the single contrast instances of the MAG framework and observed that the normal node-subgraph contrast had better detection performance than the node-node, sugraph-subgraph, and masked node-subgraph contrasts. Unlike the GRADATE\cite{gradate} model used a variety of multi-scale contrast combinations, our lightweight L-MAG surpasses the state-of-the-art on Cora and Pubmed with the low computational cost. The variant M-MAG model equipped with multi-GNN modules further improve the detection performance.
Our contributions can be summarized as follows:
\begin{itemize}
    \item To the best of our knowledge, we are the first group to unify GCAD models in the contrastive self-supervised perspective.
    \item We suggested that the multi-scale contrast modules are the \textit{"puppets"}, the backstage \textit{"pusher"} are the multi-GNN modules in GCAD.  
    \item We provided empirical design guidance for different scale contrasts and graph augmentation strategies in GCAD.
    \item The lightweight L-MAG outperforms the state-of-the-art with the low computational cost, the M-MAG improve detection performance further.
\end{itemize}

\section{Background on Graph Anomaly Detection}
For simplicity, we use capital letters, bold lowercase letters, and lowercase letters to denote matrices, vectors, and constants respectively, e.g. $X,\bm{x},x$. Giving graph $\mathcal{G}(\mathcal{V},X, A)$, $\mathcal{V}$ is composed of a series of nodes $\{\bm{v_1},\bm{v_i},...,\bm{v_n}\}$, $X\in\mathbf{R}^{n\times d}$ consists of a set of vectors $\{\bm{x_1},\bm{x_i},...,\bm{x_n}\}$, $\bm{x_i}\in\mathbf{R}^{d}$. $A\in \mathbf{R}^{n\times n}$ is the adjacency matrix of $\mathcal{G}$, where the entry $A_{i,j}$ equals to 1 if there is an edge between the $\bm{v_i}$ and $\bm{v_j}$, otherwise 0. For semi-supervised setting, we denote $\bm{y_L}=\left\{y_{l1},y_{l2},..,y_{lp}\right\}$ as the known labels, and $\bm{y_U}=\left\{y_{u1},y_{u2},..,y_{uq}\right\}$ represents the unknown labels that we have to deduce. In this section, we will brief typical graph anomaly detection techniques and formulate them below.
\subsection{GNN-based}
GNN-based methods treat anomaly detection as an unbalance binary classification task. Like GNN classifications\cite{BWGNN,fraudre,mul_gad}, we obtain node representations $Z$ via GNN mapping function $\mathcal{F}$, where $\mathcal{F}$ can be the skeleton of GCN\cite{gcn}, GAT\cite{gat} et al. The probability score $\hat{\bm{y}}\in R^{n}$ can be obtained by transforming $Z$ to $\hat{\bm{y}}$ with the multilayer perceptron. The weighted binary cross entropy between the real labeled $\bm{y_L}$ and the probability score $\hat{\bm{y}}$ will be optimized for model training.
\begin{equation}
\begin{aligned}
\label{equation_GNNbased}
    &\hat{\bm{y}} = MLP(\mathcal{F}(X,A))\\
    \mathcal{L} = \frac{1}{p}\cdot \sum_{i}^p (\alpha \cdot y_{li}&\log \hat{y_{li}}+(1-y_{li})\log (1-\hat{y_{li}}))
\end{aligned}
\end{equation}
\noindent where $p$ is the number of the known labels $\bm{y_L}$, $\alpha$ is the balance factor to regulate the imbalance between the normal and abnormal nodes. In the stage of inference, node $\bm{v_i}$ can be classified by the corresponding probability score $\hat{\bm{y}}_i$. Larger $\hat{\bm{y}}_i$, more abnormal. Based on the above procedures, Tang et.al \cite{BWGNN} analyzed from graph spectral perspective and designed spectral and spatially localized bandpass filters to better fit the anomaly detection task. Zhang et al.\cite{fraudre} concated intermediate representations, introduced fraud-aware and imbalance-oriented classification modules to overcome graph inconsistency and imbalance drawbacks in fraud detection.
\subsection{Reconstruction-based}
Reconstruction-base methods reconstruct the original graph $\mathcal{G}(\mathcal{V}, X, A)$ via the graph autoencoder architecture\cite{vgae}. It has been observed that normal nodes tend to have richer consistency with neighbouring nodes\cite{node_similarity}. Thus, normal nodes are more easily in recovering than abnormal nodes. We can identify abnormal nodes via computing the similarity between $\mathcal{G}(\mathcal{V}, X, A)$ and the reconstructed graph $\hat{\mathcal{G}}(\hat{\mathcal{V}}, \hat{X}, \hat{A})$. The forward propagation can be formulated as follows:
\begin{equation}
    \begin{aligned}
        \label{equation_reconstructedbased}
        Z = Enco&der(X,A)\\
        \hat{X} = Decoder(Z), \quad & \hat{A} = \sigma(Z*Z^T)\\
        \mathcal{L} = ||\hat{X} - X||_2& + \alpha \cdot ||\hat{A} - A||_2
    \end{aligned}
\end{equation}
\noindent where $Encoder(\cdot), Decoder(\cdot)$ can be classical GNN, such as graph convolutional network (GCN \cite{gcn}), $\alpha$ is a balance factor to regulate the errors between structure and attribute. $\sigma(\cdot)$ is sigmoid activation function to compress A to $[0,1]^{n\times n}$. In the inference, the anomaly score of node $\bm{v_i}$ can be computed by $||\hat{X_i} - X_i||_2+\alpha \cdot ||\hat{A_i} - A_i||_2$. To overcome the issues of network sparsity and label scarcity, DOMINANT \cite{dominant} as one of the classical reconstruction algorithms was presented, whose reconstructed process is just as formalized above. Differently, Fan et.al \cite{anomalydae} suggested that existing methods neglected the complex cross-modality interactions between network structure and node attribute. To this end, AnomalyDAE incorporates the attention mechanism to assess the significance of neighboring nodes, while also utilizing a dual autoencoder to enhance cross-modality representation capabilities.
\subsection{Contrastive-based}
\label{subsec: contrastive-based}
\begin{figure}[!t]
    \centering
    \includegraphics[width=0.48\textwidth]{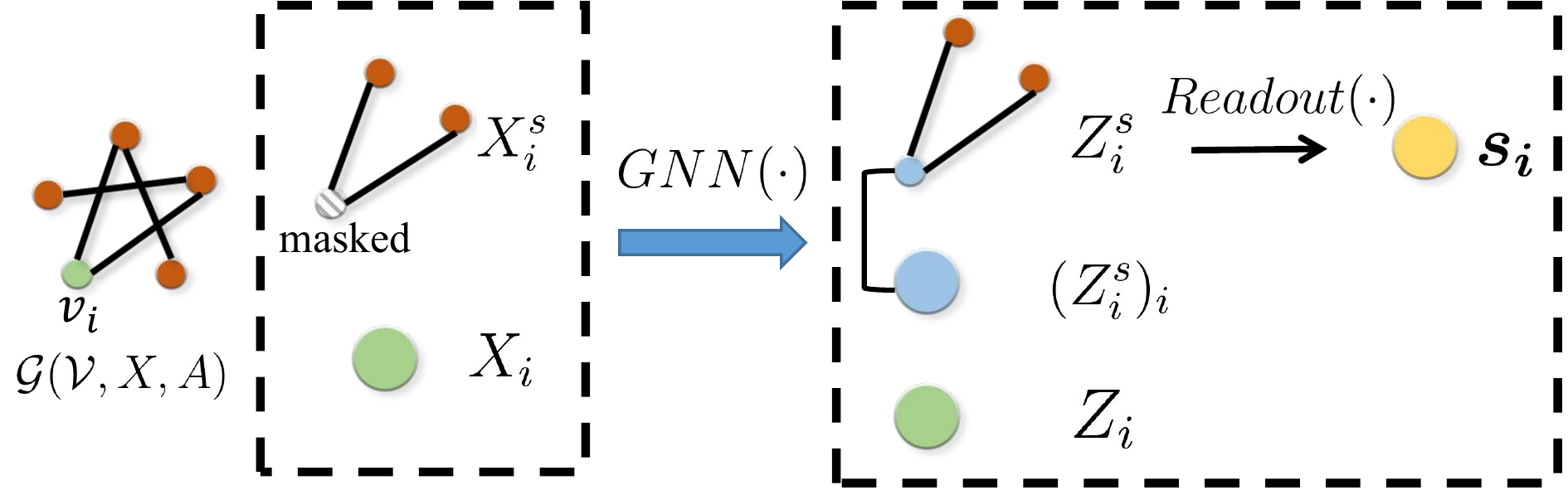}
    \caption{The three types representation of the leftmost light green node $v_i$, node feature $Z_i$ (the rightmost green), masked node feature $(Z_i^s)_i$ (the rightmost blue) and subgraph feature $\bm{s_i}$ (the rightmost yellow).}
    \label{fig:three_type}
    \vspace{-0.4cm}
\end{figure}
One of crucial modules for contrastive learning is to construct instance pairs. In GCAD, for a given node $\bm{v_i}$, we sample its subgraph $\mathcal{G}_i$ ($X_i$ in $\mathcal{G}_i$ masked with 0) using random walk restart (RWR \cite{random_walk}) method and find a distinct node $\bm{v_j},(i\neq j)$ as the negative pair of $\bm{v_i}$, where $X_i$ is the attribute features of node $\bm{v_i}$. We put the node feature $X_i\in \mathcal{R}^d$ and its sampled subgraph $\mathcal{G}_i(X^s_i,A^s_i)$ to the GNN mapping $\mathcal{F}$ to get the represented node feature $Z_i$ and its subgraph representation $Z^s_i \in \mathcal{R}^{m\times d}$, where $m$ is the number of nodes in sampled subgraph $\mathcal{G}_i$. We apply readout function to flatten $Z^s_i$ to $\bm{s_i}\in \mathcal{R}^d$. Due to $\bm{s_i}$ derived from node $\bm{v_i}$, the logical distance between $Z_i$ and $\bm{s_i}$ shall be close. Similarly, $\bm{s_j}$ derived from $\bm{v_j}$, which shall be far away from $Z_i$. We can formulate as follows:
\begin{equation}
    \begin{aligned}
        \label{equation_contrastive1}
        \bm{s_i} = Rea&dout(\mathcal{F}(\mathcal{G}_i(X^s_i,A^s_i))), \quad Z_i= \mathcal{F}(X_i)\\
        y_i = Bili&near(\bm{s_i},Z_i),\quad \hat{y_i} = Bilinear(\bm{s_j},Z_i)\\
        \mathcal{L}_1& = \frac{1}{n}\cdot \sum_i^n\log y_i+\log(1-\hat{y_i})
    \end{aligned}
\end{equation}
\noindent where $Bilinear(\cdot)$ is the bilinear function to obtain consistency score between two vectors, $n$ denotes the number of nodes. One of the classical GCAD models CoLA \cite{cola} achieve single scale node-subgraph contrast, which operates similar with the above formula. However, Jin et.al. \cite{anemone} illustrated that existing efforts only model the instance pairs in a single scale aspect, thus limiting in capturing complex anomalous patterns. To this end, ANEMONE equipped with the additional node-node contrast was proposed. Following the above expression, ANEMONE can be formulated as below.
\begin{equation}
    \begin{aligned}
        \label{equation_contrastive2}
        &Z^s_i = \mathcal{F}(\mathcal{G}_i(X^s_i,A^s_i))\\
        y^{(1)}_i = Bilinear((&Z^s_i)_i, Z_i),\quad \hat{y_i}^{(1)} = Bilinear((Z^s_j)_j,Z_i)\\
        \mathcal{L}_2 = \frac{1}{n}&\cdot \sum_i^n\log y^{(1)}_i+\log(1-\hat{y_i}^{(1)})\\
        &\quad \quad \mathcal{L} = \mathcal{L}_1 + \mathcal{L}_2
    \end{aligned}
\end{equation}
\noindent where $(Z^s_i)_i$ is $\bm{v_i}$ corresponding node feature in $Z^s_i$ as shown in Fig. \ref{fig:three_type}. Instead of contrasting with the subgraph features $\bm{s_i}$, we use $(Z^s_i)_i$ to construct positive pairs $((Z^s_i)_i,Z_i)$ and negative pairs $((Z^s_j)_j,Z_i)$ in node-node scale. Due to subgraph $\mathcal{G}_i$ masked $\bm{v_i}$ feature with 0, $(Z^s_i)_i$ can be treated as the masked node feature of $\bm{v_i}$, which have high consistency with $Z_i$. We supposed that each graph $\mathcal{G}$ can generate three type views of node $\bm{v_i}$, \textbf{subgraph features $\bm{s_i}$, node features $Z_i$, and masked node features $(Z^s_i)_i$}, as shown in Fig. \ref{fig:three_type}. It's natural to consider that subgraph-subgraph contrast shall be a promising idea for modeling more complex interaction patterns. GRADATE \cite{gradate} constructed multi-scale contrasts, including node-node, subgraph-node, subgraph-subgraph.
\subsection{Ensemble Model}
An intuitive idea comes that since anomaly detection benefits from both reconstruction and contrastive methods, taking advantage of the both shall yield a better result. SL-GAD \cite{sl_gad} was composed of generative attribute regression and multi-view contrastive learning modules to capture the anomalies. Differently, Mul-GAD \cite{mul_gad} utilized redundancy reduction techniques to eliminate the harms of similar information generated by multi-view modeling, which achieve satisfactory performance in the semi-supervised setting. 

\section{Methodology}
\label{sec:methodology}
In this section, we would detail the used GNN backbone, graph augmentation, different contrast patterns, and the final MAG framework.

\subsection{Preliminary}
We formulate the specified GNN backbone used in our MAG framework, which is well-known as graph convolutional network (GCN\cite{gcn}). The message propagation of its $l$-th layers can be formulated as follows:
\begin{equation}
    \label{equation_gcn}
    x_i^{(l)} = f_{relu}\left( \sum_{v_j \in \{v_i\} \cup \mathcal{N}(v_i)} a_{i,j}W^{(l)}x_j^{(l-1)}\right)
\end{equation}
where $x_i^{(l)}$ is the $l$-th layer representation of node $v_i$ and the $\mathcal{N}(v_i)$ denotes the collection of the $v_i$ neighbors. The $a_{i,i}$ is the entry $(i,j)$ of the $\hat{A}$, $\hat{A}=D^{-\frac{1}{2}}\overline{A}D^{-\frac{1}{2}}, \overline{A}=A+I_n, D_{i,i}=\sum_{j}\overline{A}_{i,j}$. $f_{relu}(x)=max(0,x)$ is the non-linear activation function to empower the model with non-linear modeling capability.

\subsection{Graph Augmentation}
\subsubsection{Feature Augmentation}Supposing $p$ is the probability of the node attribute being masked, $\bm{m}\in{\{0,1\}}^d$ adhered to the Bernoulli distribution $\bm{m}\sim \mathcal{B}(d,1-p)$. A augmented feature $\hat{X}$ can be computed as follows.
\begin{equation}
    \begin{aligned}
        \label{feature_augmentation}
        &\hat{X_i} = X_i\odot \bm{m}, i=1,2...n\\
        &\hat{X} = concat(\hat{X_1},..., \hat{X_{n}})
    \end{aligned}
\end{equation} 
\noindent where $\odot$ denotes the element-wise product between two vectors.
\subsubsection{Structure Augmentation}The random edge perturbation \cite{gad_review1, gad_review2} is one of the typically structure augmentation methods. Assuming $p$ is the ratio of perturbed edges. We specify the $\hat{A}$ as:
\begin{equation}
    \begin{aligned}
        \label{structure_augmentation}
        \hat{A} = A\odot (1-L) + (1-A)\odot L
    \end{aligned}
\end{equation}
\noindent where $\odot$ is element-wise multiplication and $L\in \mathcal{R}^{n\times n}$ denotes a pertubation location matrix where $L_{i,j}=L_{j,i}=1$ if node $v_i$ and $v_j$ would be perturbed. In a undirected graph, the number of perturbed edges equals to the half of $\sum_{i,j}^nA_{i,j}$. The $p$ can be calculated as $\sum_{i,j}^nL_{i,j} / \sum_{i,j}^nA_{i,j}$. Besides edge perturbation, edge diffusion\cite{edge_diffusion1, mvgcl} updates the structure via generating a different topological view. We applied two frequently used edge diffusion methods in this paper, which is Personalized PageRank (PPR) and Heat Kernel (HK). Their closed-form solutions of PPR and HK can be formulated as:
\begin{equation}
    \begin{aligned}
        \label{structure_diffusion}
        \hat{A}^{(PPR)}& = \;\alpha\left(I-(1-\alpha)D^{-1/2}AD^{-1/2}\right)^{-1}\\
        &\hat{A}^{(HK)} =exp\left(tAD^{-1}-t\right)
    \end{aligned}
\end{equation}
\noindent where $\alpha$ denotes teleport probability in a random walk and $t$ is the diffusion time. $D$ is the degree matrix of adjacency matrix $A$.
\subsection{Multi-scale Contrast}
\label{subsection_mul_contrast_in_method}
Multi-scale contrast in GCAD can be abstracted as node-node, subgraph-subgraph, and node-subgraph contrasts, which focus on different interaction patterns. By summarising the previous GCAD methods\cite{cola,anemone,gradate,sl_gad}, we noticed that graph $\mathcal{G}$ can generate three type views of node $\bm{v_i}$, \textbf{subgraph features $\bm{s_i}$, node features $Z_i$, and masked node features $(Z^s_i)_i$} as shown in Fig. \ref{fig:three_type}. These basic elements are the foundations to construct different contrast combinations. Given the graph $\mathcal{G}$, we obtain them as follows:
\begin{equation}
    \begin{aligned}
        \label{equation_method_contrastive}
        Z_i^s = \mathcal{F}(\mathcal{G}&_i(X^s_i,A^s_i)), \quad Z_i= \mathcal{F}(X_i)\\
        &\bm{s_i} = Readout(Z_i^s)
    \end{aligned}
\end{equation}
\noindent where $\mathcal{F}(\cdot)$ is GNN backbone, such as GCN\cite{gcn}, GAT\cite{gat} et al. $X_i^s$ is the neighbours of the node $v_i$, where $(X^s_i)_i$ is masked with 0. Thus, $Z_i$ is derived from node $v_i$ via GNN mapping, while $\bm{s_i}$ and $(Z_i^s)_i$ derived from the neighbors of node $v_i$.
\subsubsection{Node-node Contrast}
Node features $Z_i$ and masked node features $(Z_i^s)_i$ are utilized in this part. 
\begin{equation}
    \begin{aligned}
        \label{equation_method_node_node}
        y^{}_i = Bilinear((&Z^s_i)_i, Z_i),\quad \hat{y_i}^{} = Bilinear((Z^s_j)_j,Z_i)\\
        \mathcal{L}_{nn} = &\frac{1}{n}\cdot \sum_i^n\log y^{}_i+\log(1-\hat{y_i}^{})\\
    \end{aligned}
\end{equation}
\noindent where $(Z_i^s)_i$ is the node $v_i$ corresponding representation in $Z_i^s$. $Bilinear(\cdot)$ is the bilinear function to obtain the similarity score of the two inputs. Due to $(Z^s_i)_i$ and $Z_i$ derived from the same node, their consistency score $y_i$ is high. Conversely, the consistency between $(Z_j^s)_j$ and $Z_i$ is low. Based on the intuition, we construct loss function $\mathcal{L}_{nn}$ and optimize it.
\subsubsection{Subgraph-subgraph Contrast}We increase the subgraph views of node $v_i$ by adding a new GNN mapping $\hat{\mathcal{F}}(\cdot)$. $(\bm{s_i}, \hat{\bm{s_i}})$ and $(\bm{s_j}, \hat{\bm{s_i}})$ are employed to build the positive and negative instance pairs.
\begin{equation}
    \begin{aligned}
        \label{equation_method_subgraph_subgraph}
        \bm{\hat{s}_i} = Rea&dout(\hat{\mathcal{F}}(\mathcal{G}_i(X^s_i, A^s_i)))\\
        y_i = Bilinear(\bm{s_i}&, \bm{\hat{s}_i}),\quad \hat{y_i}^{} = Bilinear(\bm{s_j},\bm{\hat{s}_i})\\
        \mathcal{L}_{ss} = &\frac{1}{n}\cdot \sum_i^n\log y^{}_i+\log(1-\hat{y_i}^{})\\
    \end{aligned}
\end{equation}
\noindent where $\bm{s_i}$, $\bm{\hat{s}_i}$ form the position instance pairs, while $\bm{s_j}$, $\bm{\hat{s}_i}$ are regarded as negative instance pairs. 
\subsubsection{Node-subgraph Contrast}There are two expressions for node-subgraph contrasts. For identification, we call \textit{normal node-subgraph contrast} if used $Z_i$, \textit{masked node-subgraph contrast} if used $(Z_i^s)_i$.  
\begin{equation}
    \begin{aligned}
        \label{equation_method_node_subgraph1}
        y_i = Bilinear(\bm{s_i}&, Z_i),\quad \hat{y_i}^{} = Bilinear(\bm{s_j}, Z_i)\\
        \mathcal{L}_{ns}^{n} = &\frac{1}{n}\cdot \sum_i^n\log y^{}_i+\log(1-\hat{y_i}^{})\\
    \end{aligned}
\end{equation}
\noindent where $Z_i$ denotes the normal node features, while $(Z_i^s)_i$ below refers to the masked feature derived only from node $v_i$ neighbours.
\begin{equation}
    \begin{aligned}
        \label{equation_method_node_subgraph2}
         y_i = Bilinear(\bm{s_i}&, (Z_i^s)_i),\quad \hat{y_i}^{} = Bilinear(\bm{s_j}, (Z_i^s)_i)\\
        \mathcal{L}_{ns}^{m} = &\frac{1}{n}\cdot \sum_i^n\log y^{}_i+\log(1-\hat{y_i}^{})\\
    \end{aligned}
\end{equation}
\subsubsection{Inference Phase}In the training, the whole networks are updated via optimizing the contrastive loss function. In the inference stage, we obtain the consistency scores of positive and negative pairs of node $v_i$, $y_i$ and $\hat{y_i}$. For the normal nodes, the predicted score of positive instance pairs $y_i$ tended to 1, while the negative pairs $\hat{y}_i$ were closed to 0. For the anomalous node, both of the $y_i$ and $\hat{y_i}$ are closed to 0.5, which means that its positive and negative pairs would be less discriminative. Thus, the anomaly score can be computed as $(\hat{y}_i-y_i)$. Following the \cite{cola, anemone, sl_gad}, we sampled $R$ rounds to obtain the mean and standard derivation for stability. The procedure can be formulated as follows:
\begin{equation}
    \begin{aligned}
        \label{equation_inference_formula}
        f_1(v_i) = &\frac{\sum_{r=1}^R(\hat{y}_{i}^{(r)} - y_{i}^{(r)})}{R}=\overline{x}\\
        f_2(v_i) = &\sqrt{\sum_{r=1}^R((\hat{y}_{i}^{(r)} - y_{i}^{(r)}) - \overline{x})^2/R} = s\\
        &\qquad f(v_i) = \overline{x} + s
    \end{aligned}
\end{equation}
\noindent where $f(v_i)$ is the final anomaly score for node $v_i$, which denotes the sum of the mean and standard derivation. $R$ is a hyper-parameter to avoid the impact of randomness. It is suitable to set to 256, which could obtain stable result and avoid large computational costs. 
\begin{figure*}[!t]
    \centering
    \includegraphics[width=\textwidth]{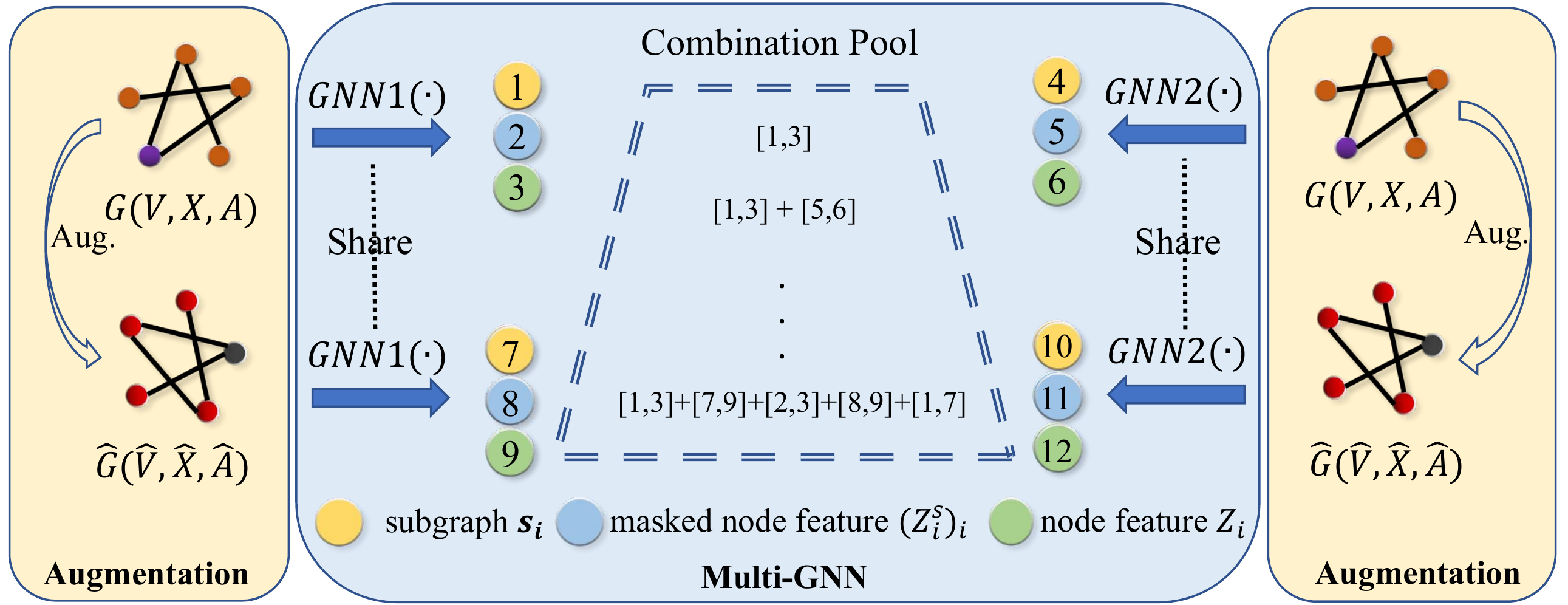}
    \caption{The overview framework of our MAG, which unified the CoLA\cite{cola}, ANEMONE\cite{anemone} and GRADATE\cite{gradate} via contrast combinations from top to bottom in the combination pool. The MAG framework consists of two modules: graph augmentation and multi-GNN modules. The normal node-subgraph, masked node-subgraph, node-node, and subgraph-subgraph contrast pairs correspond to the green-yellow, blue-yellow, green-blue, yellow-yellow pairs, respectively. For example, the [1,3]+[5,6] in the combination pool denote the used of normal node-subgraph pair and node-node pair.}
    \label{fig:method}
    \vspace{-0.4cm}
\end{figure*}
\subsection{MAG Framework}
As shown in Fig. \ref{fig:method}, each graph generates three views for node $v_i$, which is subgraph feature $\bm{s_i}$ (yellow), masked node feature $(Z_i^s)_i$ (blue), and node feature $Z_i$ (green). We increase the graph views via the graph augmentation and multi-GNN modules. The augmented graph $\hat{\mathcal{G}}$ share the training parameters with the original graph $\mathcal{G}$. The graph convolutional network (GCN\cite{gcn}) is used as GNN backbone in our framework. These views can be combined as the positive or negative instance pairs and further establish the contrastive loss function. In the combination pool, [1,3] form normal node-subgraph contrast pairs. [1,3]+[5,6] added the additional node-node contrast pairs to model complex interactive pattern. In our unified framework, different combinations are implemented by adjusting hyper-parameters, which is simple and flexible. Following the formula in section \ref{subsection_mul_contrast_in_method}, the [1,3]+[5,6] is implemented as follows:
\begin{equation}
    \begin{aligned}
        \label{equation_method_instances}
        \mathcal{L}& = \alpha \mathcal{L}_{ns}^n + \beta \mathcal{L}_{nn}\\
        f_{all}(v_i) = &\alpha f_{L_{ns}^n}(v_i) + \beta f_{L_{nn}}(v_i)  
    \end{aligned}
\end{equation}
\noindent where the $\alpha$ and $\beta$ are the balance factors to weigh different contrastive loss. In the inference stage, we obtain $f_{all}(v_i)$ as our final anomaly score. $f_{L_{ns}^n}(v_i)$ and $f_{L_{nn}}(v_i)$ can be obtained according to the formula \ref{equation_inference_formula}. The same process applied to the three or more combinations. As shown in Fig. \ref{fig:method}, the three combinations in the combination pool from top to bottom is the prototype of CoLA\cite{cola}, ANEMONE\cite{anemone}, and GRADATE\cite{gradate} methods, respectively. We compared the result of our combination with their real algorithm as shown in Table. \ref{table.unified_framework}, which show a small margin. Our MAG framework unified the classical GCAD algorithms within limited fluctuation. We further proposed the two variants of MAG, L-MAG and M-MAG. The L-MAG is the prototype of the single combination [4,9], which outperform the existing state-of-the-art on Cora and Pubmed with the low computational cost. For the multiply contrast combinations, the combination of [1,3]+[4,6] (M-MAG model) show better detection performance.
\section{Experiments and Results}
In this section, we dived into the MAG framework and provided the empirical evidence to demonstrate that our MAG model does unify the classical GCAD algorithm. To gain a deeper understanding, we propose four valuable research questions.
\begin{itemize}
\item \textbf{RQ1}: Can the MAG framework unify the classical GCAD algorithm?
\item \textbf{RQ2}: Does the graph augmentation and multi-GNN modules actually work? 
\item \textbf{RQ3}: Is the multi-scale contrast module effective?
\item \textbf{RQ4}: How is the potential of the MAG framework in single combination condition? Can the final proposed M-MAG surpass the existing methods?
\end{itemize}
\subsection{Experimental Setting}
\begin{table}[!t]
\renewcommand{\arraystretch}{1.3}
\caption{Statistics of the datasets. A half-and-half split between structure and contextual anomalies.}
\label{table.graph_data}
\centering
\setlength{\tabcolsep}{4mm}{
\begin{tabularx}{0.48\textwidth}{ccccc}
\Xhline{2\arrayrulewidth}
\textbf{Graph} & \textbf{Nodes} & \textbf{Edges} & \textbf{Features} & \textbf{Anomalies}\\
\hline
Cora & 2,708 & 5,429 & 1,433 & 150 \\
Citeseer & 3,327 & 4,732 & 3,703 & 150 \\
Pubmed & 19,717 & 88,648 & 500 & 600\\
\Xhline{2\arrayrulewidth}
\end{tabularx}}
\vspace{-1.0em}
\end{table}
\subsubsection{Datasets}Following the \cite{dgi,mvgcl,cola}, we use the three popular citation networks, Cora, Citeseer, and Pubmed\cite{cora_citeseer_pubmed}. The anomalous nodes were generated by perturbing the graph structure and modifying the node features. Thus, the graph networks are composed of structure and contextual abnormal nodes. The injection algorithm follow as \cite{cola,anemone} and the statistic detail was listed in Table. \ref{table.graph_data}
\subsubsection{Baseline}We compare with the classical shallow learning methods, Radar, and ANOMALOUS. DOMINANT and AnomalyDAE are the reconstructed-based methods. The final categories are contrastive-based methods, CoLA, ANEMONE, SL-GAD and GRADATE. For convenience, we achieve the Radar, ANOMALOUS, DOMINANT and AnomalyDAE with a python library for graph outlier detection (PyGOD\cite{pygod}). The other algorithm will be reproduced using the open source code. It is worth noting that we would set the same hyper-parameters for a fair comparison.
\subsubsection{Evaluation}The range of the anomaly score in this paper is not a probability value between [0,1]. Thus, it's not suitable to define a passing line to identify normal or anomaly nodes. The common used accuracy, precision, and recall are not taken into consider. Conversely, the Area Under Curve (AUC) is proper in this case, which will be our evaluation metrics in subsequent experiment. 
\subsubsection{Parameter Setting}
For our MAG, the training epochs and learning rate were set to 100 and 1e-3 for all datasets. The hidden dimension and batch size were set to 64, 300. We sampled 256 rounds in the inference and set the size of sampled subgraph to 4 following \cite{cola,anemone}. The balance factor was set to (0.3,0.7) for M-MAG model.

\begin{table}[!t]
\renewcommand{\arraystretch}{1.3}
\caption{The average result (AUC/\%) of CoLA, ANEMONE, and GRADATE in Cora over five seeds, compared with our corresponding MAG combination in the same hyper-parameters setting.}
\label{table.unified_framework}
\centering
\setlength{\tabcolsep}{5mm}{
\begin{tabularx}{0.48\textwidth}{ccccc}
\Xhline{2\arrayrulewidth}
\textbf{} & \textbf{CoLA} & \textbf{ANEMONE} & \textbf{GRADATE}\\
\hline
\textbf{Origin} & 89.1 & 90.6 &  90.1\\
\textbf{Our MAG} & 90.3 & 91.1 & 89.5\\
\hline
\textbf{Difference} & + 1.2 & + 0.5 & - 0.6\\
\Xhline{2\arrayrulewidth}
\end{tabularx}}
 \vspace{-1.0em}
\end{table}
\subsection{Experimental Evidence for Unified}
\begin{figure}[!t]
    \centering
    \includegraphics[width=0.48\textwidth]{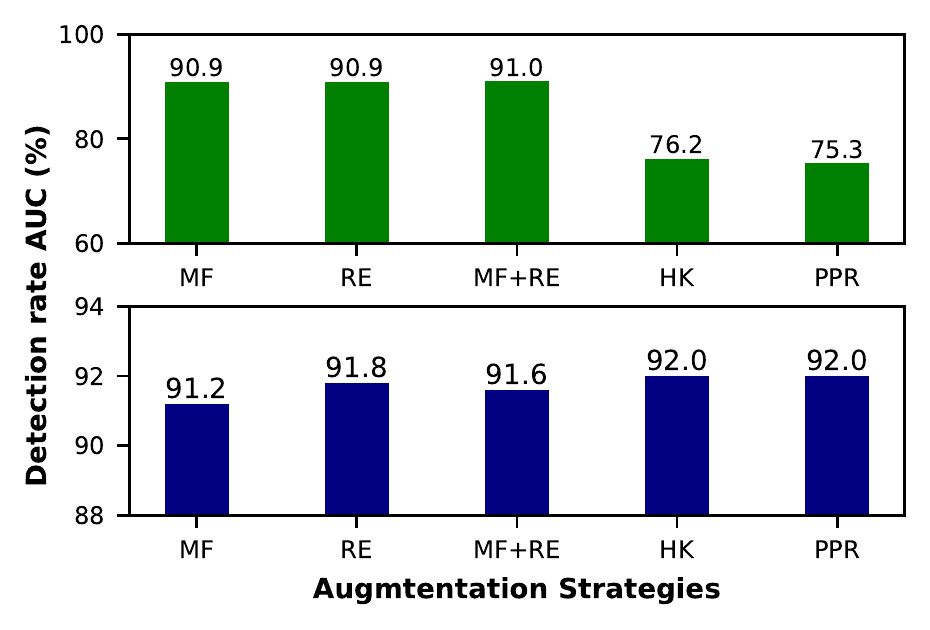}
    \caption{The MF, RE denote the masked feature, removed edge, respectively. The HK and PPR are the typical graph diffusion methods, which are heat kernel and personalized pagerank. The top one use the combination [1,3]+[7,9] of MAG framework, the bottom use [1,3]+[10,12]. Experiments are conducted on the Cora.}
    \label{fig:augmentation}
    \vspace{-0.4cm}
\end{figure}

To answer the RQ1, we compared the experimental result with CoLA, ANEMONE, and GRADATE, which are reproduced using the open source code. Their MAG combination correspond to [1,3], [1,3]+[5,6], [1,3]+[7,9]+[2,3]+[8,9]+[1,7]. Our combinations construct the similar contrast loss, while the details of the implementation are not totally same. For a fair comparison, we keep the same hyper-parameters, such as epoch, learning rate, and balance factors. For the graph augmentation module, we use the combination of masked feature and removed edge for our MAG framework, since it show a stable enhancement performance in most cases as shown in Fig. \ref{fig:augmentation}. As shown in Table. \ref{table.unified_framework}, our MAG combination models have a little margin with the corresponding algorithms, which may be caused by the different graph augmentation strategies and the randomness of the seeds. In fact, we unify GCAD model in the multi-scale contrast module, which is the most important part in GCAD. However, these relatively small margin still indicate the reasonableness of the unified model to a certain extent. Specifically, our model is highly flexible and achieve the combination of varying contrast losses by only altering the hyper-parameters.
\begin{table}[!t]
\renewcommand{\arraystretch}{1.3}
\caption{The average results (AUC/\%) of normal node-subgraph, node-node, subgraph-subgraph, masked node-subgraph contrast in Cora using corresponding single combinations of our MAG framework over five seeds, respectively.}
\label{table.different_contrast}
\centering
\setlength{\tabcolsep}{5mm}{
\begin{tabularx}{0.48\textwidth}{ccccc}
\Xhline{2\arrayrulewidth}
\textbf{} & \textbf{N-NS} & \textbf{NN} & \textbf{SS} & \textbf{M-NS}\\
\hline
\textbf{Average} & 90.96 & 86.03 & 73.81 & 69.66 \\
\hline
\end{tabularx}}
\vspace{-1.0em}
\end{table}

\begin{figure*}[!t]
    \centering
    \includegraphics[width=\textwidth]{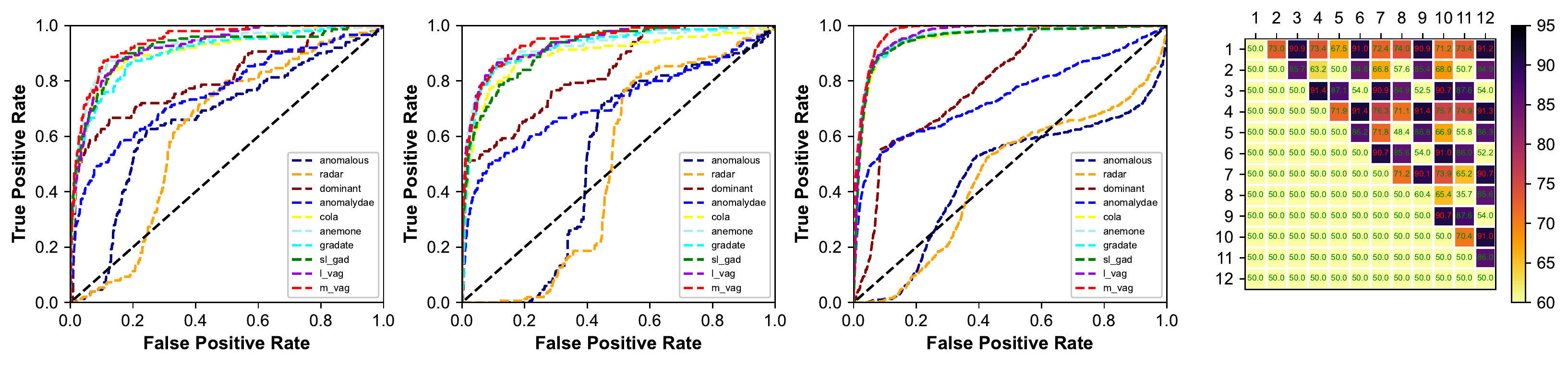}
    \caption{The three roc curves are conducted on Cora, Citeseer, and Pubmed datasets from left to right, respectively. The values in the heat plot denote the detection AUC of different contrast combination. For example, the biggest AUC value 91.4 shows repeatedly in combination [3,4], [4,6], [4,9].}
    \label{fig:roc_curve}
    \vspace{-0.4cm}
\end{figure*}
\subsection{Single Combination}
Although it is hard to traverse the search space in the multi-combination case, the single combination is feasible. To answer RQ4, we search all the single combination and plot the heat map in AUC detection rate for a clarify observation. As shown the heap map in Fig. \ref{fig:roc_curve}, the node-subgraph contrast show a excellent performance, which have a average of 90.9\%. We have summarised the other contrast patterns in Table. \ref{table.different_contrast}. The results illustrate that the ranking of gain in detection AUC by different contrast patterns are normal node-subgraph, node-node, subgraph-subgraph, masked node-subgraph, respectively. It's worth noting that one of the combinations [4, 9] even outperform the existing state-of-the-art in some situations without complex contrast combination, which would be the the lightweight instance of our MAG framework called L-MAG in subsequent experiment.
\subsection{Benefits of Graph Augmentation and Multiply GNN}
To answer the RQ2, we have compared masked feature, removed edge, masked feature + removed edge, PPR diffusion, HK diffusion for graph augmentation. The ratio of masked and removed is set to 0.2 to increase the modeling difficulties. As shown in Fig. \ref{fig:augmentation}, masked feature + removed edge have the most stable performance. We attribute the HK, PPR failures on the top bar to the limitations of the single GNN model. To examine the difference between the single and multiple GNNs, we compare the single and double GNN models. As shown in Table. \ref{table.scam}, the origin and M-G denote single and double GNNs respectively. The double one shows a higher detection AUC. We attribute the result to the fewer training parameters and the statistically unstable properties of the single GNN model.
\begin{table}[!t]
\renewcommand{\arraystretch}{1.3}
\caption{The origin, M-S, M-SG and M-G correspond to the combination [1,3], [1,3]+[2,3], [1,3]+[5,6], and [1,3]+[4,6] of the MAG framework. The best detection AUC (\%) is in bold and the runner-up is in underline, conducted on Cora over five seeds.}
\label{table.scam}
\centering
\setlength{\tabcolsep}{4mm}{
\begin{tabularx}{0.48\textwidth}{cccccc}
\Xhline{2\arrayrulewidth}
\textbf{} & \textbf{Origin} & \textbf{M-S} & \textbf{M-SG}& \textbf{M-G}\\
\hline
\textbf{Cora} & 90.3 & 90.0 &  \underline{91.1} & \textbf{91.7}\\
\textbf{Citeseer} & 91.6 & 90.0 & \underline{92.2} &  \textbf{92.5}\\
\Xhline{2\arrayrulewidth}
\end{tabularx}}
\vspace{-1.0em}
\end{table}
\subsection{Scam of Multi-scale Contrast}
We found that multi-scale modules do not improve the model performance, which multi-GNN modules do. To answer the RQ3, we constructed the origin, M-S, M-G, and M-SG as shown in Table. \ref{table.scam}, which denoted single-GNN, multi-scale, multi-GNN, and the combination of multi-scale and multi-GNN. Compared origin with M-S, the additional node-node contrast [2,3] in M-S has no benefit and even causes a corrupt performance. However, the additional node-node contrast [5,6] for M-SG get a better result. In fact, we found that the gains for M-SG derived from multi-GNN modules, not the additional node-node contrast. The result that M-G outperform the M-SG further confirm the statement. Actually, the prototype of the M-SG is the ANEMONE \cite{anemone} algorithm, which claimed that their improvement is benefited from the additional node-node contrast. They illustrated that the extra node-node contrast was able to model complex interaction patterns, which resulted the better performance. It's a scam of multi-scale modules, the multi-GNN is the hidden pushers.
\begin{table}[!t]
\renewcommand{\arraystretch}{1.6}
\caption{Comparison with the existing state-of-the-art. The detecting results AUC(\%) over five seeds, on Cora, Citeseer, Pubmed datasets. The best performance method in each experiment is in bold and the runner-up is in underline.}
\label{table.compare_with_sota}
\centering
\setlength{\tabcolsep}{5mm}{
\begin{tabularx}{0.48\textwidth}{p{2.2cm}|cccc}
\Xhline{2\arrayrulewidth}
\diagbox[]{Alg.}{Data.}& Cora& Citeseer& Pubmed\\
\hline
Radar\cite{radar}&64.8&	62.2&54.5\\
ANOMALOUS\cite{anomalous}&67.8&66.4& 54.1\\
\hline
DOMINANT\cite{dominant}&81.0&83.1&	80.5\\
AnomalDAE\cite{anomalydae}&76.2&	72.1&	78.8\\
\hline
CoLA\cite{cola}&89.1&90.6&	95.1\\
ANEMONE\cite{anemone}&90.8&91.8&	95.4\\
SL-GAD\cite{sl_gad}&91.3&91.7&	95.6&\\
GRADATE\cite{gradate}&90.1&	\underline{92.3}&94.8\\
\hline
L-MAG (Ours)&	\underline{91.4}&	91.8&\underline{95.7}\\
M-MAG (Ours)&	\textbf{91.7}&	\textbf{92.5}&\textbf{96.6}\\
\Xhline{2\arrayrulewidth}
\end{tabularx}}
\vspace{-0.3cm}
\end{table}
\subsection{Comparison with Existing Methods}
To answer the RQ4, we propose two variant models in GCAD field, L-MAG and M-MAG. L-MAG is the combination of [4,9] and M-MAG is the combination of [1,3]+[4,6] As shown in Table. \ref{table.compare_with_sota} and Fig. \ref{fig:roc_curve}, our L-MAG outperform the existing model on Cora and Pubmed with a low computational cost, while the M-MAG model further improves detection AUC benefited from the multi-GNN modules.
\section{Conclusion}
In this paper, we proposed the multi-GNN and augmented GCAD framework MAG. Our MAG framework is able to unify the classical GCAD methods by combining different contrast patterns. The proposed lightweight variant L-MAG outperform the state-of-the-art on Cora and Pubmed with the low computational cost. The variant M-MAG equipped with multi-GNN modules further improve the detection performance. Revisiting the multi-scale contrast and multi-GNN modules, we observed that the ANENONE method benefited from the multi-GNN modules, not the additional node-node contrast. We suggested that multi-scale contrast modules were the surfaced \textit{"puppet"}, while the multi-GNN modules were the real \textit{"pushers"} for complex interaction modeling. For augmentation, the masked feature and removed edge are relatively better options. In the single combination of the MAG, the normal node-subgraph express higher detection AUC than node-node, subgraph-subgraph, and masked node-subgraph contrast. The MAG framework has a vast amount of combinations, which are challenging to traverse thoroughly. Therefore, finding a better contrast combinations is a worthwhile subject. Transferring MAG framework to a more realistic scene (e.g. heterogeneous or dynamic graph) also deserves more attention.


\bibliography{ecai}

\begin{thebibliography}{10}

\bibitem{hard_to_label}
Varun Chandola, Arindam Banerjee, and Vipin Kumar, `Anomaly detection: A
  survey', {\em ACM computing surveys (CSUR)}, {\bf 41}(3),  1--58, (2009).

\bibitem{dominant}
Kaize Ding, Jundong Li, Rohit Bhanushali, and Huan Liu, `Deep anomaly detection
  on attributed networks', in {\em Proceedings of the 2019 SIAM International
  Conference on Data Mining}, pp. 594--602. SIAM, (2019).

\bibitem{gradate}
Jingcan Duan, Siwei Wang, Pei Zhang, En~Zhu, Jingtao Hu, Hu~Jin, Yue Liu, and
  Zhibin Dong, `Graph anomaly detection via multi-scale contrastive learning
  networks with augmented view', {\em arXiv preprint arXiv:2212.00535}, (2022).

\bibitem{anomalydae}
Haoyi Fan, Fengbin Zhang, and Zuoyong Li, `Anomalydae: Dual autoencoder for
  anomaly detection on attributed networks', in {\em ICASSP 2020-2020 IEEE
  International Conference on Acoustics, Speech and Signal Processing
  (ICASSP)}, pp. 5685--5689. IEEE, (2020).

\bibitem{mvgcl}
Kaveh Hassani and Amir~Hosein Khasahmadi, `Contrastive multi-view
  representation learning on graphs', in {\em International conference on
  machine learning}, pp. 4116--4126. PMLR, (2020).

\bibitem{anemone}
Ming Jin, Yixin Liu, Yu~Zheng, Lianhua Chi, Yuan-Fang Li, and Shirui Pan,
  `Anemone: Graph anomaly detection with multi-scale contrastive learning', in
  {\em Proceedings of the 30th ACM International Conference on Information \&
  Knowledge Management}, pp. 3122--3126, (2021).

\bibitem{edge_diffusion1}
Zekarias~T Kefato and Sarunas Girdzijauskas, `Self-supervised graph neural
  networks without explicit negative sampling', {\em arXiv preprint
  arXiv:2103.14958}, (2021).

\bibitem{gcn}
Thomas~N Kipf and Max Welling, `Semi-supervised classification with graph
  convolutional networks', {\em arXiv preprint arXiv:1609.02907}, (2016).

\bibitem{vgae}
Thomas~N Kipf and Max Welling, `Variational graph auto-encoders', {\em arXiv
  preprint arXiv:1611.07308}, (2016).

\bibitem{radar}
Jundong Li, Harsh Dani, Xia Hu, and Huan Liu, `Radar: Residual analysis for
  anomaly detection in attributed networks.', in {\em IJCAI}, volume~17, pp.
  2152--2158, (2017).

\bibitem{network_intrusions2}
Zhida Li, Ana Laura~Gonzalez Rios, and Ljiljana Trajkovi{\'c}, `Machine
  learning for detecting anomalies and intrusions in communication networks',
  {\em IEEE Journal on Selected Areas in Communications}, {\bf 39}(7),
  2254--2264, (2021).

\bibitem{pygod}
Kay Liu, Yingtong Dou, Yue Zhao, Xueying Ding, Xiyang Hu, Ruitong Zhang, Kaize
  Ding, Canyu Chen, Hao Peng, Kai Shu, George~H. Chen, Zhihao Jia, and
  Philip~S. Yu, `Pygod: A python library for graph outlier detection', {\em
  arXiv preprint arXiv:2204.12095}, (2022).

\bibitem{cola}
Yixin Liu, Zhao Li, Shirui Pan, Chen Gong, Chuan Zhou, and George Karypis,
  `Anomaly detection on attributed networks via contrastive self-supervised
  learning', {\em IEEE transactions on neural networks and learning systems},
  {\bf 33}(6),  2378--2392, (2021).

\bibitem{mul_gad}
Zhiyuan Liu, Chunjie Cao, and Jingzhang Sun, `Mul-gad: a semi-supervised graph
  anomaly detection framework via aggregating multi-view information', {\em
  arXiv preprint arXiv:2212.05478}, (2022).

\bibitem{money_laundering2}
Wai~Weng Lo, Siamak Layeghy, and Marius Portmann, `Inspection-l: Practical
  gnn-based money laundering detection system for bitcoin', {\em arXiv preprint
  arXiv:2203.10465}, (2022).

\bibitem{anomalous}
Zhen Peng, Minnan Luo, Jundong Li, Huan Liu, Qinghua Zheng, et~al., `Anomalous:
  A joint modeling approach for anomaly detection on attributed networks.', in
  {\em IJCAI}, pp. 3513--3519, (2018).

\bibitem{fraud_detection1}
Tahereh Pourhabibi, Kok-Leong Ong, Booi~H Kam, and Yee~Ling Boo, `Fraud
  detection: A systematic literature review of graph-based anomaly detection
  approaches', {\em Decision Support Systems}, {\bf 133},  113303, (2020).

\bibitem{cora_citeseer_pubmed}
Prithviraj Sen, Galileo Namata, Mustafa Bilgic, Lise Getoor, Brian Galligher,
  and Tina Eliassi-Rad, `Collective classification in network data', {\em AI
  magazine}, {\bf 29}(3),  93--93, (2008).

\bibitem{node_similarity}
Chaoming Song, Shlomo Havlin, and Hernan~A Makse, `Self-similarity of complex
  networks', {\em Nature}, {\bf 433}(7024),  392--395, (2005).

\bibitem{BWGNN}
Jianheng Tang, Jiajin Li, Ziqi Gao, and Jia Li, `Rethinking graph neural
  networks for anomaly detection', in {\em International Conference on Machine
  Learning}, pp. 21076--21089. PMLR, (2022).

\bibitem{money_laundering1}
Milind Tiwari, Adrian Gepp, and Kuldeep Kumar, `A review of money laundering
  literature: the state of research in key areas', {\em Pacific Accounting
  Review}, (2020).

\bibitem{random_walk}
Hanghang Tong, Christos Faloutsos, and Jia-Yu Pan, `Fast random walk with
  restart and its applications', in {\em Sixth international conference on data
  mining (ICDM'06)}, pp. 613--622. IEEE, (2006).

\bibitem{gat}
Petar Veli{\v{c}}kovi{\'c}, Guillem Cucurull, Arantxa Casanova, Adriana Romero,
  Pietro Lio, and Yoshua Bengio, `Graph attention networks', {\em arXiv
  preprint arXiv:1710.10903}, (2017).

\bibitem{dgi}
Petar Velickovic, William Fedus, William~L Hamilton, Pietro Li{\`o}, Yoshua
  Bengio, and R~Devon Hjelm, `Deep graph infomax.', {\em ICLR (Poster)}, {\bf
  2}(3), ~4, (2019).

\bibitem{gad_review1}
Lirong Wu, Haitao Lin, Cheng Tan, Zhangyang Gao, and Stan~Z Li,
  `Self-supervised learning on graphs: Contrastive, generative, or predictive',
  {\em IEEE Transactions on Knowledge and Data Engineering}, (2021).

\bibitem{gad_review2}
Yaochen Xie, Zhao Xu, Jingtun Zhang, Zhengyang Wang, and Shuiwang Ji,
  `Self-supervised learning of graph neural networks: A unified review', {\em
  IEEE transactions on pattern analysis and machine intelligence}, (2022).

\bibitem{network_intrusions1}
Vinod Yegneswaran, Paul Barford, and Johannes Ullrich, `Internet intrusions:
  Global characteristics and prevalence', {\em ACM SIGMETRICS Performance
  Evaluation Review}, {\bf 31}(1),  138--147, (2003).

\bibitem{fraudre}
Ge~Zhang, Jia Wu, Jian Yang, Amin Beheshti, Shan Xue, Chuan Zhou, and Quan~Z
  Sheng, `Fraudre: Fraud detection dual-resistant to graph inconsistency and
  imbalance', in {\em 2021 IEEE International Conference on Data Mining
  (ICDM)}, pp. 867--876. IEEE, (2021).

\bibitem{sl_gad}
Yu~Zheng, Ming Jin, Yixin Liu, Lianhua Chi, Khoa~T Phan, and Yi-Ping~Phoebe
  Chen, `Generative and contrastive self-supervised learning for graph anomaly
  detection', {\em IEEE Transactions on Knowledge and Data Engineering},
  (2021).

\end{thebibliography}
\end{document}